\documentclass[sigconf]{acmart}

\usepackage{booktabs} 
\usepackage{hyperref} 

\setcopyright{none}

\acmConference[DSHealth]{KDD Workshop on Applied Data Science for Healthcare}{August 2019}{Anchorage, AK}
\acmYear{2019}
\copyrightyear{2019}
\makeatletter
\def\runningfoot{\def\@runningfoot{}}
\def\firstfoot{\def\@firstfoot{}}
\makeatother

\begin{document}

\title{Evaluation of Embeddings of Laboratory Test Codes for Patients at a Cancer Center}

\author{Lorenzo A. Rossi, Chad Shawber, Janet Munu}

\affiliation{%
  \institution{Center for Informatics\\
  City of Hope National Medical Center}
  \city{Duarte, CA}
  \state{USA}
}
\email{lrossi@coh.org, cshawber@coh.org, jmunu@coh.org}

\author{Finly Zachariah}

\affiliation{
  \institution{Department of Clinical Supportive Care\\
  City of Hope National Medical Center}
  \city{Duarte, CA}
  \state{USA}
}
\email{fzachariah@coh.org}

\renewcommand{\shortauthors}{L.A. Rossi et al.}
\begin{abstract}
Laboratory test results are an important and generally high dimensional component of a patient's Electronic Health Record (EHR). We train embedding representations (via Word2Vec and GloVe) for LOINC codes of laboratory tests from the EHRs of about 80,000 patients at a cancer center. To include information about lab test outcomes, we also train embeddings on the concatenation of a LOINC code with a symbol indicating normality or abnormality of the result. We observe several clinically meaningful similarities among LOINC embeddings trained over our data. For the embeddings of the concatenation of LOINCs with abnormality codes, we evaluate the performance for mortality prediction tasks and the ability to preserve ordinality properties: {\em i.e.} a lab test with normal outcome should be more similar to an abnormal one than to the a very abnormal one.
\end{abstract}

\maketitle

\section{Introduction}
A patient's Electronic Health Record (EHR) contains textual data (in form of provider notes), numerical data ({\em e.g.} lab results and vitals) and sets of codes for laboratory tests, diagnoses and medications. The codes for the laboratory tests are expressed via the Logical Observation Identifiers Names and Codes (LOINC) standard. There are 70,000 LOINC codes. Especially due to the presence of free text and of different types clinical codes, EHR data requires potentially very high dimensional representations of patient information, with a challenge to design machine learning models that for many institutions like ours can only be trained over a relatively limited number of instances. 

Word embeddings techniques such as Word2Vec \citep{Mikolov2013DistributedRO} and GloVe \citep{Pennington2014GloveGV} are unsupervised approaches to represent text in low dimensional spaces. They are based on the principle that different words in similar contexts may have similar meanings and therefore can be represented by similar vectors. 
There is a growing interest in applying embeddings to the healthcare domain to generate patient representations. In \citep{choi2016learning}, the authors trained Word2Vec embeddings over LOINC, ICD-9 and NDC codes associated to the insurance claims of 4 millions of patients and evaluated them over a series of known relationships among medical concepts (e.g disease A is treated by medication B). In \citep{choi2016medical} embeddings of the aforementioned types of codes are use to predict heart failure. To date, the embeddings of medical concepts trained on the largest amount of data (insurance claims and medical notes from 60 million patients) are those presented in \citep{beam2018clinical}. Embeddings are trained from physician notes of 100,000 and 250,000 general hospital patients and evaluated over predictive tasks in  \citep{dubois2017learning}, \citep{wang2018comparison} respectively. In contrast with the settings of the aforementioned works, in our organization we have access to data from fewer patients (in the order of tens of thousands), mostly affected by cancer.

We focus on embedding representations for laboratory test data. In an analogy with the application of embedding representations to natural language processing, we can see each lab code as a word, a lab order as a sentence and a visit as a document. In our data warehouse, a patient's laboratory test record includes a LOINC code to identify the type of lab ordered and an abnormality code to indicate whether the lab result is normal or abnormal with respect to a reference lab value. We train embeddings to represent LOINC codes and the concatenations of LOINC and abnormality codes. We evaluate the lab code embeddings qualitatively by looking at their groupings in bidimensional t-SNE plots \citep{maaten2008visualizing} (Ssec. \ref{ssec:umap}). Moreover, we evaluate if certain ordinality properties of the lab outcomes are preserved in the embedding domain (Ssec. \ref{ssec:ordinality}). We aggregate the lab code embeddings to form feature representations of patients over certain time intervals and evaluate those for mortality prediction (Ssec. \ref{ssec:prediction}). 

To the best of our knowledge, this is the first work evaluating embeddings trained on clinical codes and outcome information. Our qualitative and quantitative evaluations suggest that:
\begin{enumerate}
    \item embeddings of codes associated to the same category of labs (e.g. respiratory tests) have higher similarity to each other;
    \item GloVe is better than Word2Vec at preserving ordinality properties of lab results;
    \item embeddings of the concatenation of LOINC and abnormality codes outperform bag of words representations and embeddings of just LOINC codes in mortality prediction tasks.
\end{enumerate}
We share the Word2Vec embeddings of the LOINCs and the Python code used to generate the t-SNE plot (Figure \ref{fig:tsne}) on a GitHub repository\footnote{\href{https://github.com/elleros/DSHealth2019_loinc_embeddings}
{github.com/elleros/DSHealth2019\_loinc\_embeddings}}.
We believe that this work may be useful to other healthcare institutions that are building predictive models from the EHR without access to massive amounts of data.

\section{Cohorts}
\subsection{Embedding Training Cohort}
We consider the laboratory tests ordered for a population of 79,081 inpatients and outpatients, for a total of 802,238 visits and 8,280,820 lab orders, over a period of 8 years between 2010 and 2017. This yields to a dictionary of 1098 LOINCs, after the removal of codes occurring less than 5 times. For every lab test record, there is also available an ``Abnormality Code'' with 8 possible values to indicate whether and how a result is abnormal with respect to a range of reference values for that type of patient. Such ranges were determined by the staff of the labs in our organization. Possible values of the abnormality codes are: `N', `A', `AA', `L', `LL', `H', `HH', `U', standing for normal, abnormal, extremely abnormal, low, very low, high, very high and unknown, respectively. We consider the concatenation of a LOINC code with its corresponding abnormality code as a whole word ({\em e.g.} 777-3\_N would indicate a normal platelet count). In this case, the dictionary size of the combined LOINC and abnormality codes, after removing symbols appearing less than 5 times, is 2260 (so just about 2.1 times the number of distinct LOINC codes in our data warehouse).

\subsection{Predictive Evaluation Cohort}
\begin{figure}[htbp]
  \centering 
  \includegraphics[width=\columnwidth]{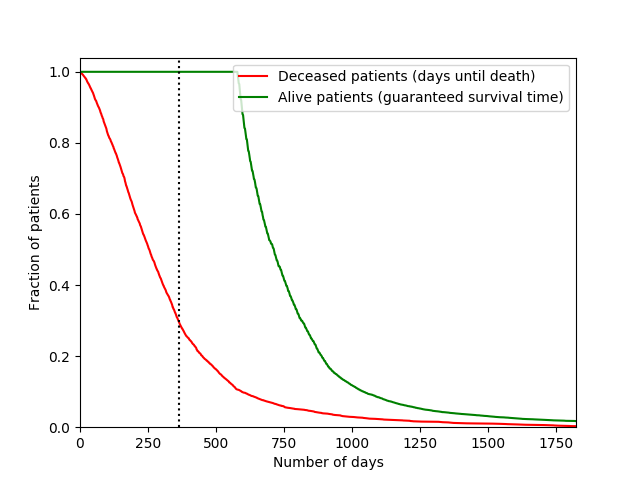} 
  \caption{Survival curves of deceased and alive patient populations.}
  \label{fig:surv} 
\end{figure}

To evaluate the predictive performance of the embeddings, we consider a cohort of about 21,000 inpatients and outpatients. 
For each patient's EHR, a \textit{prediction date} was selected from her/his encounters. For deceased patients, the prediction dates are set to achieve roughly a uniform distribution of the temporal intervals to the date of death. This implies a slow decaying profile of the survival curve. For instance, a cohort with a significant portion of prediction dates within 30 days from the dates of death would not reflect a realistic clinical situation. An alive patient was defined as not having a recorded date of death and having at least one recorded encounter 1 year after the chosen prediction date. The survival curves of deceased and alive patients are shown in Figure \ref{fig:surv}. About 3\% of the patients died within 90 days since their own prediction date.  The median ages of the alive and deceased patients are 64 and 67 years, respectively. The sample includes patients affected by solid tumors (69\%) and hematology malignancies (28\%). For each patient, we consider an \textit{observation time} of 30 days before the prediction time. LOINCs associated to lab test recorded during the observation times are aggregated to form feature representations.

\section{Methods}
We trained Word2Vec \citep{Mikolov2013DistributedRO} and GloVe \citep{Pennington2014GloveGV} embeddings from LOINC codes grouped by visit and order. Codes within the same order were shuffled. The shuffling is suggested in \citep{choi2016learning} and motivated by the fact that the ordering of LOINC codes within a given lab test order is irrelevant, unlike the case of words within a sentence. For Word2Vec embeddings we consider skip-gram and continuous bag of words (CBOW) approaches. We produced set of embeddings with dimensions from 50 to 300.

Patients' feature representations are obtained by averaging the laboratory test embeddings associated to the labs for the visits during a certain period of time. Of course, additional types of data such as demographics, diagnoses, vitals would improve the prediction performance. However, our focus is the evaluation of the lab test features alone.  The test set consists in about 3,500 records of patients whose observation windows start from March $22^{nd}, 2017$. The remaining 17,500 samples are used for cross-validation and training. Classification is performed via Logistic Regression with randomnized hyperparameter tuning. The embeddings for the prediction tasks have been trained only with records prior the above date, so from a population of of about 70,000 patients, to simulate making predictions and applying embeddings to unseen data. The embeddings for qualitative evaluation and ordinality tests are trained on the aforementioned population of 79,081 patients.

The implementations are written in Python, using open sources libraries such as pandas, Gensim \citep{rehurek_lrec}, seaborn and scikit-learn.

\section{Results} 

\subsection{Qualitative Evaluation}\label{ssec:umap}
\begin{figure}[htbp]
  \centering 
  \includegraphics[width=\columnwidth]{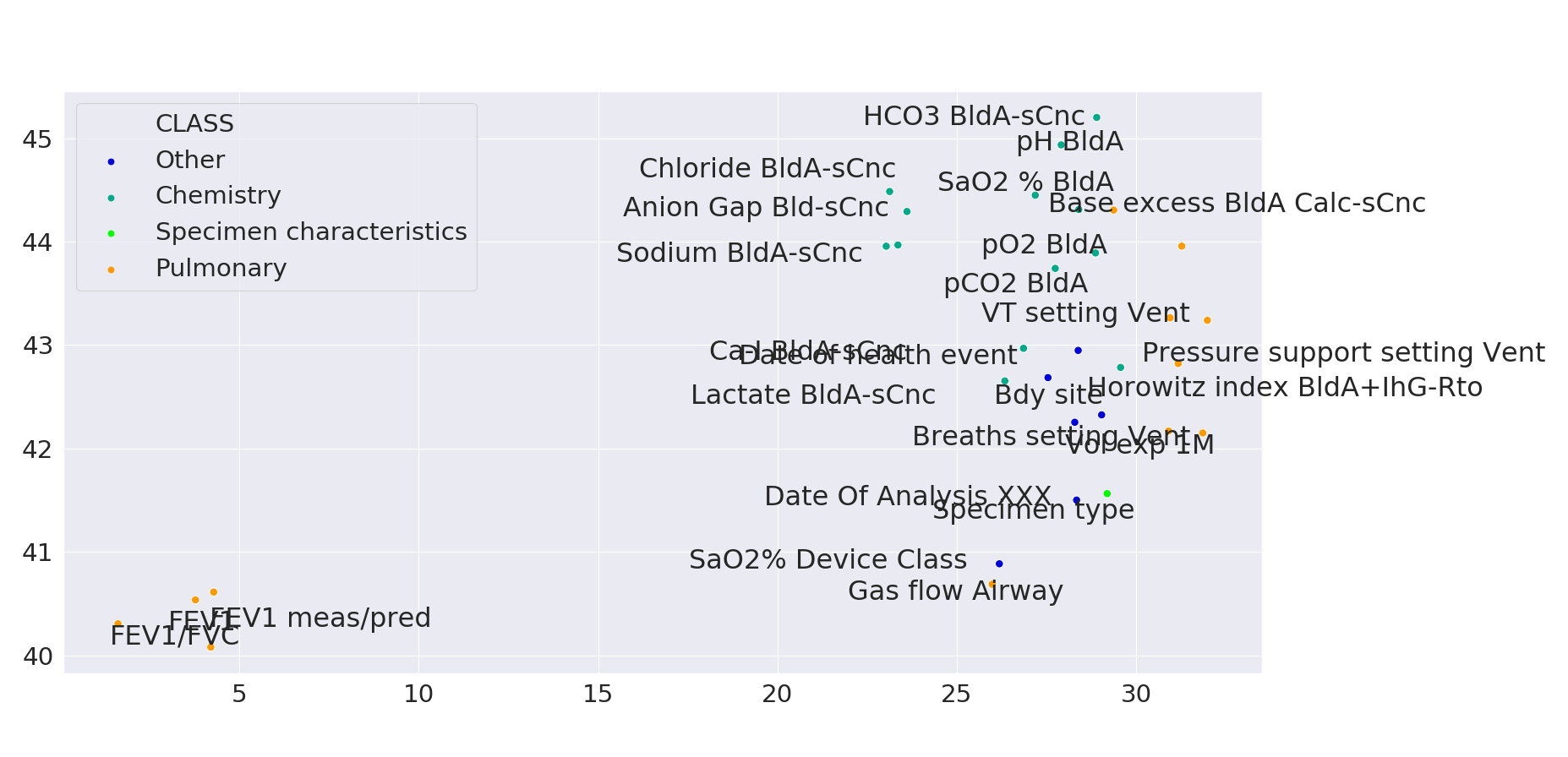} 
  \caption{t-SNE plot for cluster of respiratory lab test (enlarged from cluster at the bottom of Fig. \ref{fig:tsne}).}
  \label{fig:resp} 
\end{figure}

\begin{figure*}[htb]
  \centering 
  \includegraphics[width=\textwidth]{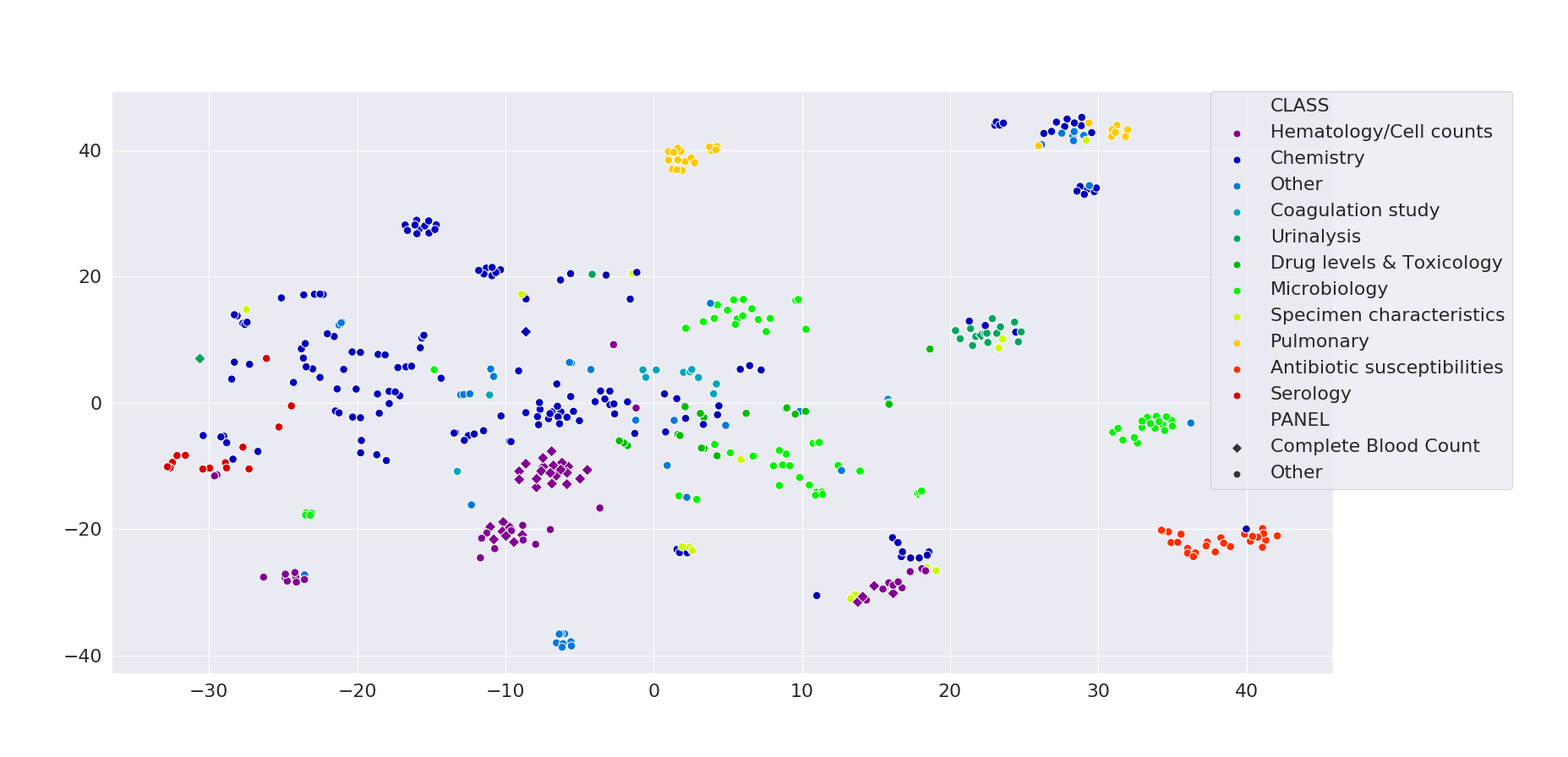}
  \caption{t-SNE plot of 500 LOINC embeddings trained via Word2Vec. The dimension of the embeddings is 200. The colors represent classes of laboratory tests defined on the LOINC.org website.}
  \label{fig:tsne} 
\end{figure*}

For the qualitative evaluation we perform t-SNE dimensionality reduction and look at groups of contiguous codes in a bidimensional plane. Figure \ref{fig:tsne} shows the t-SNE plot for the 500 most frequent LOINC embeddings trained via 200-dimensional Word2Vec (Skip-gram) vectors with size of context window equal to 5. The codes are represented in different colors according to classes of lab tests ({\em e.g.} Urynanalysis, Pulmonary). The classes of lab tests are defined in the 'LOINC Table' CSV file downloaded from the {\href{https://LOINC.org}{LOINC.org}} website. We display explicitly the 10 largest classes of labs in our dataset and group the remaining ones into the class 'Others'. Several clinically meaningful subclusters of embeddings can be noticed, {\em e.g.}: tests for respiratory capacity, antibiotic susceptibility and infectious diseases. We also highlighted the labs for Complete Blood Count panel. 
We highlight one of the clusters in Figure \ref{fig:resp}, where chemical tests for arterial blood gas are next to tests for pulmonary capacity. We provide the 200 dimensional Word2Vec embeddings along with the code to generate the plots in the repository:
{\small \href{https://github.com/elleros/DSHealth2019_loinc_embeddings}{github.com/elleros/DSHealth2019\_loinc\_embeddings}}. 

For comparison, we also trained embeddings of diagnoses expressed with ICD-9 codes (generally much less frequent than lab orders in a EHR) without observing clinically meaningful similarities. This suggests that clinically meaningful embedding representations depend on the quantity of training samples.  

\subsection{Ordinality Tests}\label{ssec:ordinality}
We use similarity measures to check whether quantitative relationships of the lab results are reflected also in the embedding models for the concatenation of LOINC and abnormality codes. For example, given a lab test with normal, abnormal and very abnormal results, we would like normal test embeddings to have higher similarity to the abnormal test ones than to the very abnormal ones, {\em i.e.} given a LOINC $code^{(k)}$ with possible test outcomes `N', `A' and `AA':
$$
S(code^{(k)}\_N, code^{(k)}\_A) > S(code^{(k)}\_N, code^{(k)}\_AA),
$$
where $S(.,.)$ is a similarity measure (e.g. cosine similarity). We defined 68 binary tests similar to the one above on our LOINC-abnormality code emebeddings. GloVe embeddings, with an error rate between 7\% and 13\%, outperform Word2Vec embeddings, whose error rate is around 50\%. Since this is a binary problem, a 50\% error rate means that no ordinality relationship is preserved with Word2Vec embeddings, at least for our training data. This does not exclude that with a larger training set, Word2Vec could have preserved the ordinality properties. 

\subsection{Predictive Evaluation}\label{ssec:prediction}
\begin{figure*}[htb]
  \centering 
    \includegraphics[width=\textwidth]{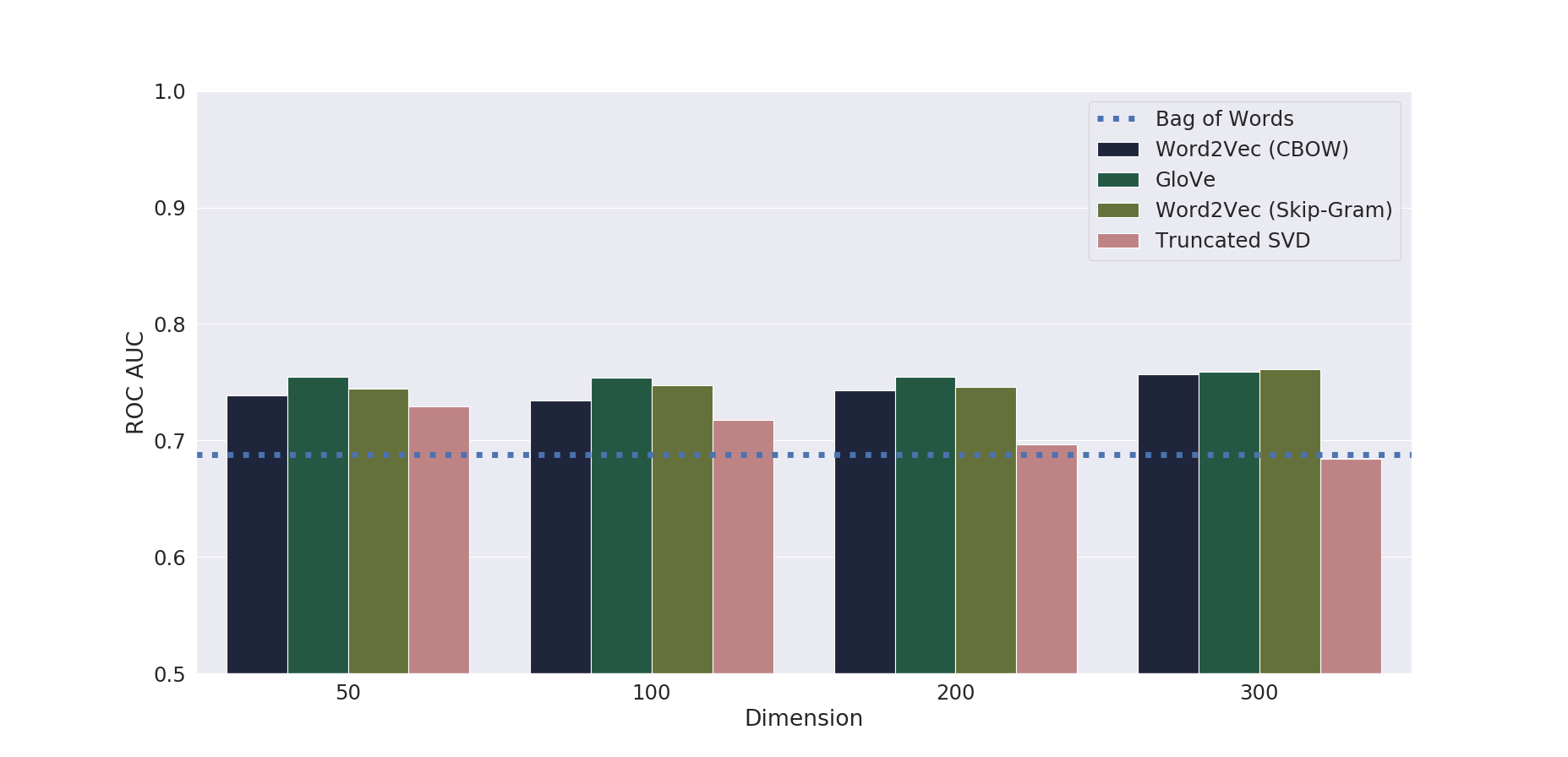}
    \caption{Comparison of ROC AUC for features derived from embeddings in a 90 day mortality prediction task. The dimension of the representations is on the horizontal axis. Truncated Singular Value Decomposition (SVD) is a dimensionality reduction technique applied to the bag of words representation. Continuous Bag of Words (CBOW) is an approach to train Word2Vec embeddings.}
    \label{fig:d90_prediction} 
\end{figure*}

For the quantitative evaluation, we consider the task of predicting 90 days patient mortality. Mortality prediction is very relevant to a cancer center population, because it enables better decisions to optimize the end of life experience of the patients, {\em e.g.} to recommend clinical trials with the right timing or to avoid overtreatment. 
We compare the results with the bag of word (BOW) set of features and its truncated singular value decomposition representation (which is equivalent to PCA, but more suitable for sparse data). The resulting feature vectors are used to train and test logistic regression classifiers. To compare performances, we use the area under the receiver operating characteristic curve (ROC AUC).\footnote{If our goal were to have a full assessment of the performance, we should also include the precision-recall curve, since the labels are highly imbalanced.}. Results are shown in Figure \ref{fig:d90_prediction}, where the ROC AUCs for BOW, truncated SVD and embeddings are compared. For the embeddings, we considered dimensions from 50 to 300 and context windows equal to 5. Overall, embeddings perform better than the BOW and truncated SVD baselines with an AUC ranging from 0.74 to 0.76. 300 dimensional embeddings generally outperformed lower dimensional embeddings. GloVe and skip-gram perform slightly better than continuous bag of words. The ordinality preservation property of the GloVe embeddings (Ssec \ref{ssec:ordinality}) does not seem to lead to a superior prediction performance. We also varied the duration of the observation window ({\em e.g.} to 60 days) and tried embedding aggregations such as median, min and/or max, but with inferior results.

Besides, we evaluated embeddings of LOINC codes without bnormality symbols, obtaining AUC scores below 0.65. 
To the best of our knowledge, the only publicly available set of embeddings encompassing LOINC codes is the one presented in \citep{choi2016learning}.\footnote{\href{https://github.com/clinicalml/embeddings}{github.com/clinicalml/embeddings}} The embeddings are trained over sets of LOINC, ICD-9 and NDC codes from 4 million insured subjects. The LOINCs in this embedding model are about 3,500, with 493 of them in common with our dataset. We compared the prediction performance restricted to the aforementioned 493 embeddings. Overall the  AUC score from the public embeddings trained over insurance claims is slightly below the performance of the embeddings trained over our patient records. 

\section{Conclusion}
We have trained embeddings for laboratory test data, in form of LOINC and concatenations of LOINC and abnormality codes, for a population of 79,081 cancer patients. The embeddings have been trained via the skip-gram, CBOW and GloVe approaches for different dimensions and sizes of the context window. 
The embeddings of LOINC + abnormality codes yield to better prediction results than the embeddings of just LOINC codes providing a potentially useful tool for low dimensional representations of lab test outcome. Generally, embedding based features outperform in prediction tasks features from baselines such as bag of words and truncated SVD. Moreover, GloVe embeddings seem to preserve ordinality properties of the lab test outcomes such as the pair (abnormal, normal) having higher similarity than (very abnormal, normal).


\bibliographystyle{ACM-Reference-Format}
\bibliography{clinical_ml}

\end{document}